\documentclass[review,3p,times]{elsarticle}
\usepackage{amssymb}
\usepackage{graphicx}
\usepackage{amsmath}
\usepackage{amssymb}
\usepackage{booktabs}
\usepackage{hyperref}
\usepackage{verbatim}
\usepackage{enumitem} %
\usepackage{tabularx}
\usepackage{amsmath}
\usepackage{multirow}
\usepackage{tabularx}
\usepackage{array} 
\usepackage{amssymb}
\journal{Applied Energy}

\begin{document}

\begin{frontmatter}
\begin{highlights}
\item Proposes a novel framework for evaluating the power generation potential of building facades.
\item Employs a vision-language foundation model to achieve pixel-accurate segmentation of facade components, requiring no domain-specific training.
\item Structured prompt chains guide Large Language Models (LLMs) in spatial reasoning to convert semantic masks into code-compliant layouts.
\item A homographic transformation guided by facade semantic keypoints eliminates keystone distortion in street-level imagery.
\end{highlights}

\title{Solar PV Installation Potential Assessment on Building Facades Based on Vision and Language Foundation Models}
\begin{abstract}
Building facades represent a significant untapped resource for solar energy generation in dense urban environments, yet assessing their photovoltaic (PV) potential remains challenging due to complex geometries and semantic components. This study introduces SF-SPA (Semantic Facade Solar-PV Assessment), an automated framework that transforms street-view photographs into quantitative PV deployment assessments. The approach combines computer vision and artificial intelligence techniques to address three key challenges: perspective distortion correction, semantic understanding of facade elements, and spatial reasoning for PV layout optimization. Our four-stage pipeline processes images through geometric rectification, zero-shot semantic segmentation, Large Language Model (LLM)-guided spatial reasoning, and energy simulation. Validation across 80 buildings in four countries demonstrates robust performance with mean area estimation errors of 6.2\% $\pm$ 2.8\% compared to expert annotations. The automated assessment requires approximately 100 seconds per building, a substantial gain in efficiency over manual methods. Simulated energy yield predictions confirm the method’s reliability and applicability for regional potential studies, urban energy planning, and building-integrated photovoltaic (BIPV) deployment. Code is available at: \url{https: github.com/CodeAXu/Solar-PV-Installation}.

\end{abstract}

\author[dtu]{Ruyu Liu}
\ead{ruyli@dtu.dk}
\author[tjut]{Dongxu Zhuang}
\author[tjut]{Jianhua Zhang}
\author[dtu]{Arega Getaneh Abate}
\author[dtu]{Per Sieverts Nielsen}
\author[hznu]{Ben Wang}
\author[dtu]{Xiufeng Liu\corref{cor1}}

\cortext[cor1]{Corresponding Author}

\affiliation[dtu]{
    organization={Technical University of Denmark},
    department={Department of Technology, Management and Economics},
    city={Lyngby},
    country={Denmark}
}

\affiliation[hznu]{
    organization={Hangzhou Normal University},
    department={School of Computer Science and Technology},
    city={Hangzhou},
    country={China}
}

\affiliation[tjut]{
    organization={Tianjin University of Technology},
    department={School of Computer Science and Engineering},
    city={Tianjin},
    country={China}
}


\end{frontmatter}
\section{Introduction}

Despite recent advancements, most building-integrated photovoltaic (BIPV) potential studies focus predominantly on rooftop installations\cite{Kevin}. This rooftop-centric approach overlooks the significant energy generation and architectural integration potential of vertical façades, particularly in dense urban areas with limited roof space and increasing building heights. Neglecting façade PV potential can lead to a significant underestimation of a city's total renewable energy capacity\cite{Abdalrahman, Zhiling, ZhilingGuo}. However, assessing façade PV potential presents unique challenges. Unlike near-planar roofs, which are easily surveyed with aerial imagery, façades demand a different approach. As complex vertical structures, façades require observation from multiple viewpoints for a comprehensive analysis. Furthermore, common data sources like street-level imagery suffer from perspective distortion. Without proper correction, these distortions introduce significant errors into geometric measurements and solar irradiance models\cite{Kechuan}.

Building façades also exhibit a complex hierarchy of semantic components. These components include opaque walls (potential PV sites), windows (requiring semi-transparent PV or exclusion), balconies, doors, and various decorative or shading elements. Existing assessment methods often simplify façades into uniform planes, focusing on geometry while ignoring these critical semantic details. This oversight hinders accurate PV technology selection, layout optimization, and energy yield prediction. Studies have shown that methods ignoring semantic details typically overestimate installable capacity by 25-40\% compared to detailed architectural analysis, leading to significant economic miscalculations in BIPV project planning \cite{Zhe, Qingyu}.

To address these limitations, we propose the Semantic Façade Solar-PV Assessment (SF-SPA) pipeline, a novel framework for unlocking the untapped potential of façade-integrated solar resources. This automated framework is designed to convert a single, readily available street-level photograph of a building into robust, quantitative PV deployment metrics and energy yield estimates. Our workflow proceeds in four main stages. First, it computes a homography from semantic keypoints (e.g., window corners) to rectify perspective distortion. Second, a vision-language model performs zero-shot semantic segmentation on the rectified image to delineate all architectural elements. Third, a Large Language Model (LLM) applies codified design rules---such as edge clearances and module constraints---to the resulting semantic masks to generate a precise map of the installable area. Finally, this map's attributes are input to the \texttt{pvlib} toolkit to simulate irradiance and estimate the annual energy yield. This chained pipeline provides an automated, scalable, and semantically aware workflow capable of supplanting manual CAD-based studies and enabling efficient, city-scale screening of vertical solar potential. The framework assesses gross solar potential, providing an upper-bound estimate based on available area and clear-sky irradiance.

The primary contribution is a novel, automated pipeline that translates a single 2D image into a metrically-accurate, installable PV layout and energy estimate. This process uniquely chains geometric rectification, zero-shot semantic parsing, and LLM-based spatial reasoning, eliminating the need for 3D data, manual intervention, or domain-specific models. The specific contributions of this paper are:
\begin{itemize}[leftmargin=*]
    \item \textbf{Automated Façade PV Assessment Framework:} Uses vision-language models to perform semantic perception and spatial reasoning from a single street-view image, quantifying installable area and predicting energy yield.
    \item \textbf{Distortion-Aware Rectification:} Employs semantic keypoints to perform a homographic transformation, correcting perspective distortion in street-level images.
    \item \textbf{Zero-Shot Façade Parsing:} Leverages vision-language models to segment façade components at the pixel-level, eliminating the need for specialized, annotated training datasets.
    \item \textbf{Prompt-Guided Spatial Reasoning:} Introduces a novel ``describe'' $\rightarrow$ ``partition'' $\rightarrow$ ``filter'' $\rightarrow$ ``summarise'' prompt chain to guide an LLM in converting raw segmentation into practical PV layouts.
\end{itemize}

The remainder of this paper is structured as follows: Section~\ref{sec:related_work} reviews relevant prior work; Section~\ref{sec:methodology} details the proposed pipeline; Section~\ref{sec:results} presents our experimental results; and Section~\ref{sec:conclusions} summarizes our findings and outlines future work.
\section{Related Work}
\label{sec:related_work}
Research on solar potential in the built environment, driven by computer vision and machine learning, has split into two main streams: (i) rooftop-centric and (ii) whole-building assessments.

\subsection{Rooftop-Centric PV Assessments}
Rooftops are the traditional site for BIPV installations because their direct sun exposure and simple geometry simplify modeling and installation. Early methods relied heavily on Geographic Information System (GIS) data and manual digitization of building footprints \cite{Wiginton2010}. More recent computer vision (CV) approaches use overhead or oblique remote-sensing imagery to automatically outline roof segments and detect existing PV arrays. For instance, 3D-PV-Locator \cite{Mayer2022} fuses LiDAR point clouds with aerial images to generate 3D bounding boxes for rooftop modules, enabling precise city-scale localization and orientation estimation. Hong et al. \cite{hang} used GIS-based shadow modeling and meteorological data to analyze sunlight exposure on rooftops over time, enabling a holistic assessment of PV potential. Other studies use high-resolution imagery and digital surface models (DSMs) to find suitable rooftop areas; for example, Hu et al. \cite{Hu} combined this approach with the Normalized Difference Vegetation Index (NDVI) to improve object identification in Beijing. TransPV \cite{Guo2024} employs Vision Transformers to significantly improve the accuracy of PV panel detection and segmentation on high-resolution orthophotos. SolarSAM \cite{Li2024SolarSAM} advances automation by combining the Segment Anything Model (SAM) \cite{Kirillov2023} with text-prompt engineering. This combination enables zero-shot segmentation of rooftops from satellite imagery, allowing for the estimation of installable BIPV areas without locality-specific training data. Many urban-energy studies, such as the work by Wang et al. \cite{Wang2025}, subsequently integrate these derived semantic masks (roof areas) into thermal-electric co-simulation platforms to model building energy performance.
Despite these technological advances, rooftop-centric approaches face several critical limitations that constrain their applicability to comprehensive urban energy planning. First, many algorithms operate purely in 2D space, failing to accurately recover roof dimensions, orientations, and complex geometries, which introduces significant uncertainty into installable area and energy yield estimates. Second, the exclusive focus on rooftops systematically ignores the substantial energy potential of vertical façades—a limitation that becomes increasingly problematic in dense, high-rise urban environments where façade area often exceeds rooftop area by factors of 3-5.

More fundamentally, rooftop-only assessments fail to capture the full potential of building-integrated photovoltaic systems, which require holistic building envelope analysis to optimize energy generation, architectural integration, and economic viability. This narrow focus has contributed to systematic underestimation of urban renewable energy potential and missed opportunities for comprehensive building energy optimization.

\subsection{Whole-Building Shell Assessments}
To overcome the limitations of rooftop-only studies, a second research stream analyzes the entire building shell, including façades. This work is enabled by open 3D city models (e.g., CityGML) and advances in 3D sensing technologies like LiDAR and photogrammetry \cite{Zhang2018, Miclea2021}. Typical pipelines simplify buildings into ``box-like'' geometries from existing 3D models, then ray-cast sun vectors onto their surfaces to compute annual solar irradiation for each segment. Cheng et al. \cite{Cheng2020} calculated roof and façade solar potential by extruding 2D building footprints into 3D box models. They sampled the building envelope at a 3-meter resolution, corrected irradiance with 30-year historical data, and mapped solar resources for buildings in ten major Chinese cities. Recent advancements have seen more sophisticated methodologies emerge for evaluating solar PV potential on building surfaces. 

Deep learning has significantly advanced façade parsing, improving segmentation precision and object detection in complex urban settings. Contemporary methodologies specifically address two persistent challenges: architectural occlusion and facade heterogeneity. Key methods address façade heterogeneity and occlusion. For example, Zhang et al. \cite{Zhang} used a multi-scale attention framework for semantic segmentation, while Liu et al. \cite{Liu1} integrated architectural principles into CNNs. Similarly, Zhou et al. \cite{Zhou} introduced occlusion reasoning to interpret partially obscured elements. For rural areas, Liu et al. \cite{Liu} generated 3D models from satellite imagery and GIS data to assess both rooftop and façade potential. Building on this, Chen et al. \cite{Chen} used a remote-sensing approach to estimate building heights and segment footprints from aerial imagery for a comprehensive PV potential evaluation. However, current techniques face notable limitations - while LiDAR and 3D footprint data provide geometric information, they lack critical semantic details about facade features, resulting in less precise PV assessments for vertical surfaces compared to rooftops. Furthermore, the field still predominantly considers facade PV potential as secondary to rooftop analyses, rather than treating it as an equally important component of building-integrated solar energy systems. Recent advanced pipelines show further integration. Li et al. \cite{Li2025} introduced a UAV-driven framework for façade-aware PV deployment. Dong et al. \cite{Dong2025} combined zero-shot SAM with 3D shadow simulation for element-level assessments. Finally, Yu et al. \cite{Yu2025} merged 3D building data with meteorological datasets, showing that façades significantly augment rooftop PV potential.

While these whole-building approaches represent significant progress toward comprehensive solar potential assessment, they suffer from critical methodological limitations that constrain their practical applicability. Most critically, existing methods treat façades as uniform, geometrically simple planes, fundamentally ignoring the complex semantic hierarchy of architectural components that governs real-world PV installation feasibility. This oversimplification leads to systematic overestimation of installable capacity, as methods fail to account for windows, doors, balconies, decorative elements, and structural constraints that significantly reduce usable area.

Furthermore, current approaches typically require expensive 3D data acquisition (LiDAR, photogrammetry) or rely on simplified building models that may not accurately represent complex architectural geometries. The dependence on specialized datasets and domain-specific training limits scalability and adaptability across diverse urban contexts and architectural styles.

\subsection{Research Gap and Contribution}
The literature review reveals a fundamental gap in façade PV assessment methodology: the lack of automated, scalable approaches that combine geometric accuracy with semantic understanding of architectural components. Existing methods either focus exclusively on rooftops (missing substantial façade potential) or treat façades as uniform surfaces (ignoring installation constraints). No current approach successfully integrates readily available street-view imagery with advanced AI techniques to provide both geometric rectification and semantic-aware PV layout optimization.

Our proposed SF-SPA framework addresses this gap through three key innovations: (1) semantic-guided geometric rectification that corrects perspective distortions using architectural features, (2) zero-shot façade parsing that eliminates the need for domain-specific training data, and (3) LLM-driven spatial reasoning that translates semantic understanding into practical installation layouts. This integration enables automated, scalable façade PV assessment using only street-view imagery, representing a significant advancement in building-integrated photovoltaic potential evaluation.

\section{Methodology}
\label{sec:methodology}
\subsection{Overview}
\begin{figure}
    \centering
    \includegraphics[width=0.9\columnwidth]{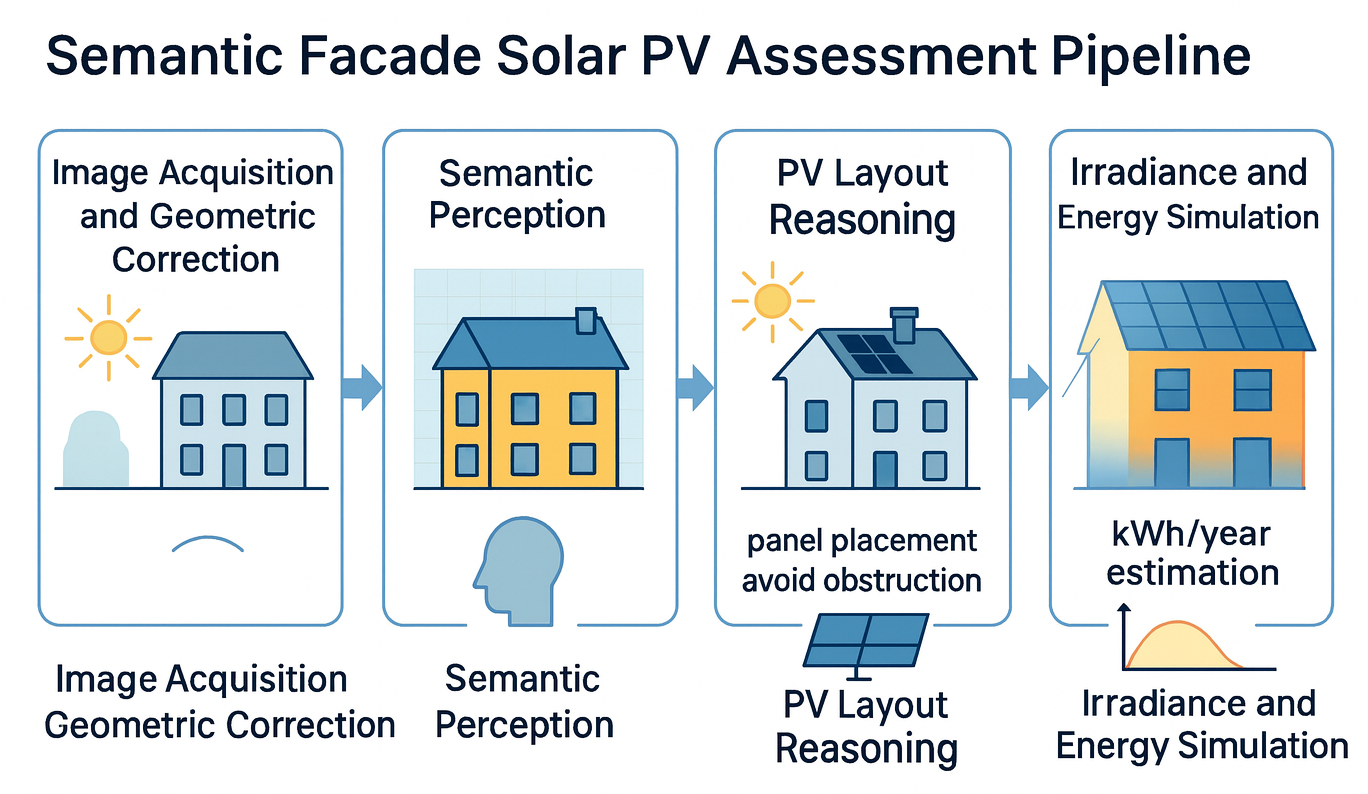} %
    \caption{The Semantic Façade Solar-PV Assessment (SF-SPA) pipeline. The process consists of four primary stages from image acquisition to energy simulation.}
    \label{fig:overall_pipeline_diagram}
\end{figure}

As illustrated in Figure \ref{fig:overall_pipeline_diagram}, the proposed Semantic Façade Solar-PV Assessment (SF-SPA) pipeline systematically converts a single, uncalibrated street-view photograph of a building façade into an actionable photovoltaic (PV) installation blueprint and its associated annual energy yield estimate. The pipeline comprises four sequential, interconnected stages: (i) Façade Image Acquisition and Geometric Rectification, (ii) Semantic Perception with a Vision Foundation Model, (iii) LLM-Driven PV Layout Reasoning, and (iv) Irradiance and Energy Simulation. Each stage builds upon the output of the preceding one, ensuring a cohesive workflow from a raw image to actionable energy intelligence.

\subsection{Facade Image Acquisition and Geometric Rectification}
Although readily available, street-view photographs are captured from arbitrary viewpoints and thus rarely align with the building façade plane. This misalignment introduces perspective distortions (e.g., the keystone effect), which can corrupt pixel-based semantic masks and subsequent metric area calculations. To correct these distortions, we employ a local, semantics-guided homography transformation consisting of three steps:

\begin{enumerate}[leftmargin=*]
    \item \textbf{Source-point selection:} Identifying a façade's global corner points from a single street-level image is often unreliable due to occlusions (e.g., trees, cars, adjacent buildings) or the façade extending beyond the image frame. Therefore, our approach focuses on detecting a local, planar semantic component. A window is typically chosen for this purpose due to its planarity, repetitive presence on façades, and relative ease of segmentation. The four vertices of the selected window in the source image serve as the source points, $p = \{(x_i, y_i, 1)\}_{i=1}^4$, for homography estimation.

    \item \textbf{Target-point initialisation and homography solving:} Since the window's true metric dimensions are unknown a priori, its corresponding target points, $p'$, are initialized in the rectified view as the vertices of an axis-aligned unit square. Alternatively, a rectangle with a canonical aspect ratio can be used if prior knowledge of window shapes is available, although a square is generally robust for enforcing perpendicularity. This choice enforces the perpendicularity and parallelism of the window edges in the rectified view. Consequently, the keystone skew is removed without requiring information about global façade corners or camera intrinsic parameters. The $3 \times 3$ homography matrix $H$ is then computed by solving the system of equations derived from $p' \cong Hp$. This is typically done using Direct Linear Transform (DLT) \cite{DLT} followed by non-linear refinement, minimizing the geometric error $||p' - Hp||$. Robust estimation techniques like Random Sample Consensus (RANSAC) \cite{RANSAC} or Least Median of Squares (LMedS) \cite{LMedS}, available in libraries such as OpenCV (\texttt{cv2.findHomography}), are employed to handle potential inaccuracies in source point selection. The computed homography $H$ is then applied to the entire input image using \texttt{cv2.warpPerspective} to obtain a fronto-parallel view of the façade segment that contains the reference window. The mapping relationship between any point $q = [x, y, 1]^T$ in the source image and its corresponding point $q' = [x', y', 1]^T$ in the target view is defined by:
    \begin{equation}
    \begin{vmatrix}
     x^{'} \ y^{'} \ 1^{}
    \end{vmatrix}
    \cong H
    \begin{vmatrix}
      x \ y \ 1
    \end{vmatrix}
          =
    \begin{vmatrix}
     h_{11}& h_{12}&h_{13}\\ h_{21}&h_{22}&h_{23}\\ h_{31}&h_{32}&h_{33}  
    \end{vmatrix}
    \begin{vmatrix}
      x \ y \ 1
    \end{vmatrix}
    \end{equation}

    \item \textbf{Metric scaling:} Although the homography transformation corrects perspective distortion, the rectified image is still expressed in pixel units and lacks a physical scale. To establish this metric scale, we use widely available online mapping services (e.g.,  \href{https://www.google.com/maps}{Google Maps},    \href{https://www.openstreetmap.org}{OpenStreetMap}) that often provide tools for measuring real-world distances. Using such a tool, we manually measure the approximate real-world width of the target building facade ($w_m$) in metres. In the rectified, fronto-parallel image, we then measure the corresponding facade width in pixels ($w_{px}$), ensuring the measured segment aligns with the real-world measurement. The scale factor, $s$, is then computed as $s = w_m / w_{px}$ (in metres per pixel). This scale factor, s, enables the conversion of any pixel-based measurements from the rectified image, such as the areas of segmented regions or dimensions of potential PV layouts, into metric units (m, m$^2$). This step is essential for all subsequent PV layout optimization and energy calculations.

    \textbf{Sensitivity Analysis for Manual Scaling:} We acknowledge that the manual scaling step introduces potential error propagation into the final area and energy estimates. Since area calculations are proportional to the square of the linear scale factor, errors in width measurement have a quadratic impact on area estimates. To quantify this sensitivity, we conducted a systematic analysis on a representative subset of 20 buildings from our dataset. We introduced controlled scaling errors of ±5\%, ±10\%, ±15\%, and ±20\% to the manually measured facade width and propagated these through the entire pipeline. The results show that a ±10\% error in width measurement leads to approximately ±20\% error in area estimation and ±19\% error in energy yield prediction. To mitigate this limitation, we recommend: (1) using multiple reference measurements when possible (e.g., door heights, window dimensions), (2) cross-validation with publicly available building footprint data, and (3) employing semi-automated approaches such as detecting standard architectural elements with known dimensions (e.g., standard door height $\approx$ 2.1m). Future work will explore fully automated scaling using stereo vision or integration with 3D city models.
\end{enumerate}

\subsection{Semantic Perception with a Vision Foundation Model}
Conventional façade-PV studies that move beyond coarse geometric analysis typically require either laborious manual annotation of façade elements or the training of specialized models to segment each component of interest (e.g., windows, walls, doors). This reliance on bespoke, annotated datasets creates a significant bottleneck, limiting the scalability and adaptability of such methods to diverse architectural styles. Inspired by recent advances in zero-shot learning driven by VFMs \cite{Dong2025, Li2024SolarSAM}, we adopt a large-model, zero-shot segmentation strategy. This approach eliminates the need for a dedicated, façade-annotated training dataset and significantly expands the range of detectable components.

\begin{table}[t!]
\centering
\caption{Model results on the ODinW benchmark for Full-Shot Setting.}
\label{tab:Grounding-Dino}

\begin{tabular}{cccc}
\toprule
\textbf{Model} & \textbf{Language Input} & \textbf{Model Size} & \textbf{Test(APaverage)} \\
\midrule
GLIP \cite{GLIP}           & \multicolumn{1}{c}{$\surd$} & 232M & 62.6 \\
DyHead-T \cite{DyHead}       & \multicolumn{1}{c}{$\times$}  & 100M & 63.2 \\
DINO-Swin-T \cite{DINO-Swin}    & \multicolumn{1}{c}{$\times$}  & 49M  & 66.7 \\
OmDet \cite{OmDet}        & \multicolumn{1}{c}{$\surd$} & 230M & 67.1 \\
DINO-Swin-L \cite{DINO-Swin}    & \multicolumn{1}{c}{$\times$}  & 218M & 68.8 \\
Grounding DINO \cite{Liu2024} & \multicolumn{1}{c}{$\surd$} & 172M & 70.7 \\
\bottomrule
\end{tabular}%

\end{table}

We employ the open-source Lang-Segment-Anything (LSA) framework \cite{Liu2024}. LSA combines the capabilities of two foundation models:
\begin{itemize}[leftmargin=*] 
    \item \textbf{Grounding-DINO \cite{Liu2024}:} An open-set object detector that locate arbitrary objects based on natural language text prompts. As shown in Table \ref{tab:Grounding-Dino}, a comparative analysis on standard benchmarks demonstrates that Grounding-DINO achieves state-of-the-art Average Precision (AP) with fewer parameters than competing models, while accepting natural language prompts as input. \newline 
    \hspace*{2em}AP is a comprehensive metric that evaluates detection performance across various recall levels. It is particularly suitable for assessing a model's ability to recognize and accurately locate arbitrary objects. The AP is calculated as::
    \begin{align}
    AP = \displaystyle \sum_{k=1}^{N} P(k) \Delta R(k)  
    \end{align}
where $N$ is the total number of data points, $P(k)$ is the precision at the $k-th$ point, and $\Delta R(k)$  is the change in recall between point $k-1$ and $k$.

    \item \textbf{Segment Anything Model (SAM) \cite{Kirillov2023}:} A promptable segmentation model that generates high-quality object masks from various prompt types, including points, boxes, or masks supplied by other models. We use the latest version of SAM for improved performance. As shown in Table \ref{tab:SAM2}, SAM demonstrates strong segmentation performance on several public datasets (MOSE \cite{MOSE}, DAVIS 2017 \cite{DAVIS}, LVOS \cite{LVOS}, SA-V). We use the combined Jaccard and F-score (J\&F) metric for evaluation. \newline
    \hspace*{2em}The J\&F metric provides a comprehensive evaluation of segmentation performance by averaging the region similarity (J) and the boundary accuracy (F). It is calculated as:
    \begin{equation}
    J\&F = \frac{J + F}{2}
    \end{equation}
    The \textbf{Jaccard index ($J$)}, or region similarity, measures the overlap between the predicted segmentation (P) and the ground-truth segmentation (G). It is calculated as the ratio of the area of their intersection to the area of their union. A higher J value indicates better segmentation accuracy.
    \begin{equation}
    J = \frac{|P \cap G|}{|P \cup G|}
    \end{equation}
    
    Where:
    \begin{itemize}
        \item $P$ represents the predicted segmentation region
        \item $G$ represents the ground truth segmentation region
        \item $|\cdot|$ denotes the pixel count of the region
    \end{itemize}

    \hspace*{2em}The \textbf{F-score ($F$)}, or boundary accuracy, evaluates the alignment between the predicted and ground-truth boundaries. It is the harmonic mean of boundary precision and recall. A higher F value signifies a better match between the predicted and ground-truth boundaries.
    \begin{equation}
    F = \frac{2 \times \text{Precision} \times \text{Recall}}{\text{Precision} + \text{Recall}}
    \end{equation}
    
    Where:
    \begin{itemize}
        \item \textbf{Precision}: Measures what fraction of predicted boundary pixels are actual boundary pixels
        \item \textbf{Recall}: Measures what fraction of actual boundary pixels are correctly predicted as boundary pixels
    \end{itemize}
    
    Note that boundary precision and recall are typically computed with a small tolerance for minor misalignments, usually by dilating the boundary pixels before matching.
    
\end{itemize}

Our pipeline begins by providing a single natural-language prompt (e.g., “segment wall, window, door, balcony, roof, and other façade parts”) to Grounding-DINO. Grounding-DINO then processes the rectified façade image using this prompt to generate coarse bounding boxes for each identified component. These bounding boxes, in turn, serve as prompts for the SAM model, which refines them into precise, pixel-level segmentation masks. Since the entire process is prompt-driven, the system is highly extensible. New component types can be recognized simply by modifying the text prompt, which bypasses the significant cost and effort required to create large-scale, annotated datasets.

\begin{table}[]
\caption{SAM performance in video segmentation tasks based on first-frame ground-truth mask prompts.}
\label{tab:SAM2}
{\tiny
\resizebox{\columnwidth}{!}{
\begin{tabular}{cccccc}
\toprule
\multirow{3}{*}{\textbf{Method}} & \multicolumn{5}{c}{J\&F}         \\ \cline{2-6}
 & 
  \begin{tabular}[c]{@{}c@{}}MOSE\ val
  \end{tabular} & 
  \begin{tabular}[c]{@{}c@{}}DAVIS\ 2017 val
  \end{tabular} & 
  \begin{tabular}[c]{@{}c@{}}LVOS\ val
  \end{tabular} & 
  \begin{tabular}[c]{@{}c@{}}SA-V\ val
  \end{tabular} & 
  \begin{tabular}[c]{@{}c@{}}SA-V\ test
  \end{tabular} \\ \midrule
STCN \cite{STCN}       & 52.5 & 85.4 & -    & 61.0 & 62.5 \\
SwinB-DeAOT \cite{SwinB} & 59.9 & 86.2 & -    & 61.4 & 61.8 \\
ISVOS \cite{ISVOS}                  & -    & 88.2 & -    & -    & -    \\
DEVA \cite{DEVA}                   & 66.0 & 87.0 & 55.9 & 55.4 & 56.2 \\
Cutie-base \cite{Cuite}             & 69.9 & 87.9 & 66.0 & 60.7 & 62.7 \\
SAM \cite{Kirillov2023}                  & 77.9 & 90.7 & 78.0 & 77.9 & 78.4 \\
\bottomrule
\end{tabular}%
}}
\end{table}

A morphological post-processing step (e.g., closing small gaps, removing small isolated regions) is applied to the raw segmentation masks. This step merges negligible fragments with their nearest valid regions and removes spurious detections, thereby preventing the generation of impractically small or fragmented PV patches. The final set of masks provides both the geometric boundaries (pixel coordinates) and semantic labels (e.g., wall, window) for all relevant façade elements. These structured data are then forwarded to the LLM-driven layout-reasoning module described in Section \ref{sec:llm_reasoning}.

To account for systematic errors in the LSA segmentation, we introduce a bias correction step based on empirical validation. Our analysis of the 80-building validation set revealed a consistent pattern: the vision-language model tends to slightly under-segment wall areas (mean under-estimation of 3.2\%) while over-segmenting window areas (mean over-estimation of 2.1\%). This systematic bias likely stems from the model's conservative approach to boundary detection and its training on diverse architectural styles.

We validated the assumption of systematic bias through statistical analysis. The Shapiro-Wilk test confirmed that segmentation errors follow a normal distribution (p > 0.05 for all building types), and the one-sample t-test showed significant deviation from zero bias (p < 0.001), justifying the correction approach. The bias correction is calculated on a validation subset (20 buildings) and applied to the remaining test set to avoid overfitting.

The Bias term measures the mean relative error between ground-truth areas ($S$) and model-predicted areas ($\hat{S}$):

\begin{align}
Bias = \frac{ {\textstyle \sum_{1}^{n}S_{i}} - {\textstyle \sum_{1}^{n}\hat{S_{i}}} }{{\textstyle \sum_{1}^{n}S_{i}}}
\end{align}

For new predictions, the initial estimated area $\hat{S}$ is adjusted using this pre-computed bias:

\begin{align}
S = \frac{\hat{S} }{1 - Bias}
\end{align}

We acknowledge this approach assumes consistent bias across building types and image conditions. Future work will explore building-type-specific corrections or regression-based calibration methods for improved robustness.

\subsection{LLM-Driven PV Layout Reasoning}
\label{sec:llm_reasoning}
The semantic masks from the previous stage detail the identity and location of components on the façade. However, converting this information into a practical PV layout requires spatial reasoning that incorporates metric scale, semantic exclusions (e.g., no panels on windows), and design constraints (e.g., module sizes, clearances). Traditional algorithmic approaches to this problem are often complex to encode and inflexible. We use the reasoning capabilities of Large Language Models (LLMs) for this task, guided by a four-stage prompt chain. This chain mirrors the logical operations of a human designer: \textit{describe} \textrightarrow \textit{transcribe} \textrightarrow \textit{partition} \textrightarrow \textit{qualify}. Our analysis employs GPT-4 for its robust spatial reasoning and instruction-following capabilities, although the framework is adaptable to other LLMs. The full prompt template is provided in the Appendix.

\textbf{Step 1—Describe metric canvas:} The prompt first provides the LLM with the necessary context. This includes the pixel dimensions of the rectified image ($w_{px}, h_{px}$) and its corresponding real-world size in metres ($w_m, h_m$), derived using the scale factor $s$ from §3.2.3. This step anchors all subsequent geometric reasoning in a physical, metric scale, allowing the LLM to process lengths and areas in real-world units.

\textbf{Step 2—Transcribe semantic layout:} Each component mask (e.g., wall, window, door) from the LSA model is translated into a plain-text list of bounding boxes. For example: \texttt{window: [[x1\_w1, y1\_w1, x2\_w1, y2\_w1], [x1\_w2, y1\_w2, x2\_w2, y2\_w2], ...]}, \texttt{door: \\ \hspace{2em}[[x1\_d1, y1\_d1, x2\_d1, y2\_d1], ...]}. This structured transcription provides the LLM with the spatial coordinates of all obstructions and usable areas, focusing its capabilities on logical reasoning over this textual input rather than on visual inspection.

\textbf{Step 3—Partition usable wall:}
The LLM is instructed to analyze the \textquotedblleft residual wall\textquotedblright~area—the total façade area minus obstructions like windows and doors. It must subdivide this space into a set of non-overlapping rectangular regions suitable for PV installation. The instructions follow three key rules:
    (a) Maximize the size of each region: each identified rectangular region should be as large as possible, and the total number of such regions should be minimal to favor larger, contiguous PV arrays.
    (b) Ensure mutual exclusivity and complete coverage: the generated regions must not overlap with each other, nor with any of the predefined obstruction areas. Together, they should aim to cover the entire usable wall area.
    (c) Merge adjacent sub-regions: if merging two or more adjacent, smaller valid sub-regions results in a larger valid region without violating exclusivity or obstruction rules, this should be done to optimize for larger panel groups.

\textbf{Step 4—Qualify for PV installation:}
The LLM performs a final check on each candidate region from Step 3 against practical installation constraints. The primary constraints, applicable only to wall regions, are based on standard PV module dimensions: a candidate rectangle must have a short edge of at least 1.0 m and a long edge of at least 1.2 m. These default values represent a typical module size and are user-configurable. Rectangles that do not meet these minimum dimensions are discarded. The final output is a list of qualified rectangles and their total aggregated area (in m$^2$). 

\subsection{Irradiance and Energy Simulation}
\label{sec:energy_conversion}
Building on the usable area rectangles from Section \ref{sec:llm_reasoning}, this final stage converts these geometries into annual energy yield estimates. This provides a quantitative measure of the façade's solar energy potential under specific geographic and climatic conditions. This process utilizes the \texttt{pvlib} Python toolkit \cite{Holmgren2018}, a standard library for PV performance simulation.

\textbf{Irradiance modelling:}  Each identified façade requires defined geographical coordinates (latitude, longitude) and a true-north azimuth (orientation). This orientation can be estimated from map data or EXIF tags, or assumed from common building practices if the data is unavailable. A nominal 90° surface tilt is assumed for all vertical facades. These parameters, along with the date and time, are used as inputs for the irradiance models in the \texttt{pvlib} library. For an ideal upper-bound estimate, hourly Global Horizontal Irradiance (GHI), Direct Normal Irradiance (DNI), and Diffuse Horizontal Irradiance (DHI) are synthesized using a clear-sky model, such as Ineichen \cite{Ineichen2002}. For more realistic, weather-aware simulations, these irradiance components are sourced from Typical Meteorological Year (TMY) datasets (e.g., PVGIS, NREL TMY3) or historical weather records, such as ERA5 reanalysis data. The \texttt{pvlib} library then transposes these components onto the tilted façade plane to calculate the Plane of Array (POA) irradiance.
\begin{align}
    E_{POA} &= E_{dir} + E_{dif} + E_{ref} 
\end{align}
\begin{align}
   E_{dir} &= DNI \cdot \cos(AOI)  
\end{align}
\begin{align}
   E_{dif} &= DHI \cdot \frac{1 + \cos(\beta)}{2} 
\end{align}
\begin{align}
    E_{ref} &= GHI \cdot \rho \cdot \frac{1 - \cos(\beta)}{2} \end{align}

Here, DNI, DHI, and GHI are the direct normal, diffuse horizontal, and global horizontal irradiance inputs. The key parameters are the Angle of Incidence (AOI), the façade tilt angle ($\\beta$), and the ground's albedo ($\\rho$). By calculating these components hourly for each façade while accounting for the AOI, the model accurately captures temporal variations in solar exposure, thus preventing overestimations of energy availability.

To improve accuracy, the model also incorporates the effect of temperature on the performance of the Building-Integrated Photovoltaic (BIPV) system. Meteorological data (ambient temperature and wind speed) from the PVGIS-ERA5 database are used in the Sandia Array Performance Model (SAPM) to determine the module's thermal characteristics. The model computes the module back-surface temperature ($T_{m}$) and cell temperature ($T_{c}$) based on incident irradiance ($E_{POA}$), ambient temperature ($T_a$), and wind speed ($W_s$), as shown in Equations (12) and (13):

\begin{align}
T_{m} = E_{} \cdot e^{a + b \cdot Ws} + T_{a}
\end{align}

\begin{align}
T_{c} = T_{m} + \frac{E_{POA}}{E_{0}}\\triangle T
\end{align}

In these equations, $E_0$ is the reference irradiance (1000 W/m²), $\\triangle T$ is an empirical temperature difference coefficient, and $a$ and $b$ are module-specific parameters derived from the SAPM. Finally, the BIPV power output is estimated using the single-diode model in \texttt{pvlib}, with the calculated cell temperature ($T_{c}$) and POA irradiance ($E_{POA}$) as inputs.

\textbf{Temporal horizon:} The default simulation covers a full calendar year at an hourly resolution to capture seasonal variations. This temporal horizon can be adjusted—for instance, shortened for seasonal studies or lengthened for multi-year climate assessments—without altering the fundamental pipeline structure.

\textbf{System configuration:} To determine the number of installable panels, the area of each eligible rectangle (m²) is divided by the footprint of a reference PV module (e.g., 1.2$m^2$). The electro-thermal parameters for the modules (e.g., standard opaque crystalline silicon) are then sourced from \href{https://pvlib-python.readthedocs.io/}{pvlib}'s extensive component libraries. Parameters for semi-transparent BIPV glazing can also be used, provided that windows are identified as potential installation sites and the modeling rules are adjusted accordingly. Finally, a corresponding inverter model with integrated Maximum Power Point Tracking (MPPT) is selected to complete the system configuration.

\textbf{Power conversion:} The single-diode model in \texttt{pvlib} calculates the DC power output from the PV array based on the POA irradiance and module temperature. This DC power is subsequently converted to AC power using the selected inverter model. The model assumes a standard module technology (e.g., crystalline silicon at 20\% efficiency) and a corresponding inverter efficiency (e.g., 97\%). Standard system losses (e.g., soiling, wiring resistance, module mismatch, and light-induced degradation) are also incorporated using \texttt{pvlib}'s tunable default parameters. The hourly AC power profile is then integrated over the simulation period to yield the total annual energy production (kWh). This process yields a defensible, façade-specific energy estimate suitable for building energy-balance analyses or urban-scale energy planning. The core DC power calculation is based on the single-diode model, which relates the module's output current ($I$) and voltage ($V$):

\begin{align}
P = V \cdot I
\end{align}

\begin{align}
I = I_{L} - I_{0}[exp(\frac{V+IR_s}{nN_{s}V_{th}})-1] - \frac{V+IR_s}{R_{sh}}
\end{align}

Here, $I$ is module output current, $V$ is module output voltage, $I_{L}$ is photocurrent, $I_{0}$ is diode reverse saturation current, $R_{S}$ is series resistance, $R_{sh}$ is shunt resistance, $n$ is diode ideality factor, $N_{s}$ is  number of cells in series, $V_{th}$ is thermal voltage.  
\section{Experimental Results and Analysis}
\label{sec:results}
The performance, robustness, and practical utility of the proposed SF-SPA framework were evaluated across a diverse range of urban contexts. The evaluation used a custom dataset of 80 building samples (Table \ref{tab:buildings}). These samples were curated to represent diverse geographical locations (e.g., Tianjin, China; Ålesund, Norway; Ankara, Turkey) and building typologies (e.g., high-rise commercial, mid-rise residential, and low-rise industrial). Our evaluation includes quantitative metrics—installable area accuracy, energy yield deviation, and processing time—and qualitative visualisations. The visualisations overlay predicted PV layouts onto source images and 3D reconstructions, offering an intuitive assessment of the system's output. We also conducted ablation studies to quantify the contribution of each key module: geometric rectification, semantic segmentation, the LLM prompting strategy, and the \texttt{pvlib} energy simulation. Together, these experiments provide a holistic assessment of the framework's capabilities and its readiness for real-world application.

\subsection{Ground Truth Validation Protocol}
\label{sec:ground_truth}
To ensure the reliability of our performance metrics, we established a rigorous ground truth annotation protocol for the 80-building dataset. The annotation process involved three qualified experts: two licensed architects with 5+ years of experience in building-integrated photovoltaic design, and one solar energy engineer with expertise in PV system installation codes and standards. Each expert independently annotated the installable PV area for all 80 buildings using a standardized protocol.

\textbf{Annotation Tools and Software:} Experts used AutoCAD 2023 and QGIS 3.28 for precise area measurements. The rectified facade images were imported as georeferenced rasters, enabling accurate metric measurements. Each expert was provided with: (1) the original street-view image, (2) the rectified facade image with established metric scale, (3) building metadata (location, orientation, construction year), and (4) local building codes and PV installation standards.

\textbf{Annotation Guidelines:} The installable area definition followed IEC 61215 standards and local building codes, requiring: (1) minimum 0.5m clearance from building edges, (2) exclusion of areas with permanent shading (balconies, overhangs), (3) avoidance of windows, doors, and ventilation systems, (4) consideration of structural load-bearing capacity, and (5) compliance with fire safety access requirements. Experts were instructed to identify contiguous rectangular regions suitable for standard PV module installation (1.65m × 1.0m modules).

\textbf{Inter-Annotator Agreement:} To assess annotation consistency, we calculated the inter-annotator agreement using the Jaccard index across all 80 buildings. The mean Jaccard index was 0.89 ± 0.06, indicating high consistency among experts. Cases with Jaccard index < 0.80 (8 buildings) underwent consensus review, where experts discussed discrepancies and reached agreement. The final ground truth represents the consensus annotation, ensuring reliable validation of our automated framework.

\subsection{Qualitative and Quantitative Evaluation}
\begin{figure}[t!]
    \centering
    \includegraphics[width=\columnwidth]{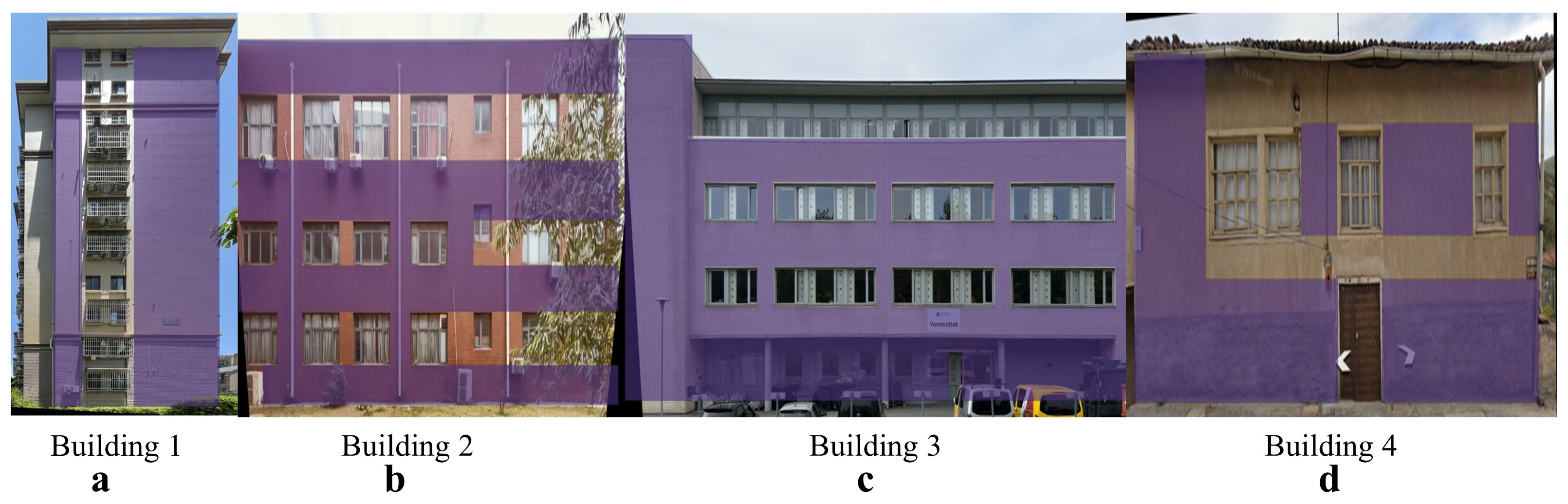} %
    \caption{Predicted PV installation patches (colored rectangles) overlaid on representative façades from the dataset: (a) Hangzhou, China; (b) Tianjin, China; (c) Ankara, Turkey; (d) Ålesund, Norway. This illustrates the method's adaptability to diverse architectural styles.}
    \label{fig:pv_patches_representative}
\end{figure}

To establish a performance baseline, we first provide a qualitative and quantitative summary of the framework. Figure \ref{fig:pv_patches_representative} overlays the LLM-identified PV installation areas (colored rectangles) onto representative façade images from our dataset. The figure illustrates the framework's adaptability to diverse architectural styles, from modern glass towers to mid-rise residential blocks and low-rise industrial buildings. The visualizations show the system populating continuous, unobstructed surfaces with PV modules while correctly avoiding features such as windows, balconies, and doors.

\begin{table}[t!]
\caption{Distribution of Building Samples.}
\label{tab:buildings}
{\tiny
\resizebox{\columnwidth}{!}{
\begin{tabular}{ccccc}
\toprule
Location        & Low-rise building(s) & Mid-rise building(s) & High-rise building(s) & Total \\
\midrule
Hangzhou, China & -                    & 10                   & 10   & 20                 \\
Tianjin, China  & -                    & 10                   & 10    & 20                \\
Ankara, Turkey  & 10                   & 10                   & -       & 20              \\
Ålesund, Norway & 10                   & 10                   & -  & 20    \\
\bottomrule
\end{tabular}% 
}}
\end{table}

Table \ref{tab:core_metrics_expanded} presents three key quantitative metrics: 1) installable PV area, 2) annual incident solar irradiation, and 3) annual AC energy yield. The area estimation error, $\\varepsilon$, is computed by comparing the framework's identified areas to an expert-annotated ground truth. Discrepancies are quantified as false negatives ($S_1$) and false positives ($S_2$). The error $\\varepsilon$ is calculated as:
\begin{equation}
    \\varepsilon = \frac{S_1 + S_2}{S}
\end{equation}
where $S$ represents the total expert-approved (ground truth) installable area. 

These metrics provide an end-to-end performance view, connecting the spatial opportunity ($m^2$) to the energy benefit (kWh), and serve as a baseline for interpreting ablation studies. Across the 80-building test set, the framework achieved an average area estimation error of 6.2\%, demonstrating high accuracy in identifying usable surfaces. Furthermore, the end-to-end processing time averaged 100 seconds per building on a standard GPU-equipped computer (Table \ref{tab:total-time}), highlighting the framework's computational efficiency.

\begin{table}[]
\caption{Core Metrics of Photovoltaic Potential for Representative Building Façades, including Expert-Defined Ground Truth Area and Estimation Error.}
\label{tab:core_metrics_expanded}
\resizebox{\columnwidth}{!}{%
\begin{tabular}{ccccccc}
\hline
\textbf{Building ID} &
  \textbf{Building Location} &
  \textbf{\begin{tabular}[c]{@{}c@{}}Expert-Defined \ Area (m$^2$)
  \end{tabular}} &
  \textbf{\begin{tabular}[c]{@{}c@{}}Estimated Installable \ Area (m$^2$)
  \end{tabular}} &
  \textbf{\begin{tabular}[c]{@{}c@{}}Area Estimation \ Error (\%)
  \end{tabular}} &
  \textbf{\begin{tabular}[c]{@{}c@{}}Annual Clear-Sky \ Irradiation (kWh/m$^2$)
  \end{tabular}} &
  \textbf{\begin{tabular}[c]{@{}c@{}}Clear-Sky \ Energy Yield (kWh)
  \end{tabular}} \\ \hline
1 & Hangzhou, China & 120.50 & 112.26 & 6.8 & 689.58  & 13,391.8 \\
2 & Tianjin, China  & 198.20 & 189.38 & 4.5 & 919.82  & 27,622.4 \\
3 & Ankara, Turkey  & 52.90  & 49.21  & 7.0 & 1356.22 & 10,667.3 \\
4 & Ålesund, Norway & 322.00 & 305.13 & 5.2 & 849.86  & 36,021.4 \\
\hline
\end{tabular}% 
}
\end{table}

\begin{table}[]
\caption{End-to-End Processing Time per Building (Image to Energy Yield)}
\label{tab:total-time}
\resizebox{\columnwidth}{!}{%
\begin{tabular}{cccc}
\toprule
 &
  \textbf{\begin{tabular}[c]{@{}c@{}}Average Semantic \ Segmentation Time(s)
  \end{tabular}} &
  \textbf{\begin{tabular}[c]{@{}c@{}}Average LLM \ Inference Latency(s)
  \end{tabular}} &
  \textbf{\begin{tabular}[c]{@{}c@{}}PV Energy Yield \ Computation Time(s)
  \end{tabular}} \\
\midrule
SF-SPA & 
  0.274s & 
  97.03s & 
  0.412s    \\
\bottomrule
\end{tabular}% 
}
\end{table}

\begin{figure}[t!]
    \centering
    \includegraphics[width=\textwidth]{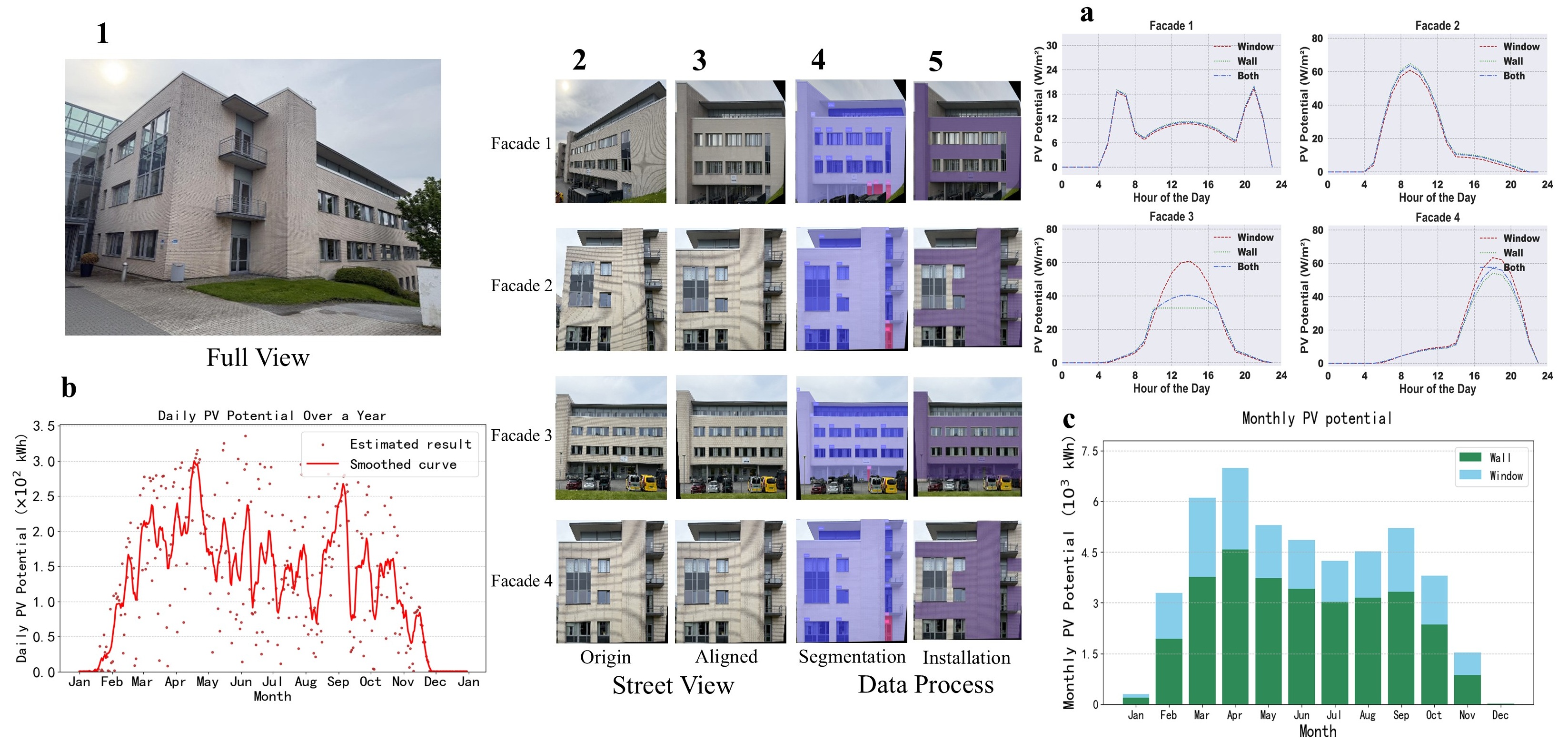} %
    \caption{End-to-end workflow demonstration for facade PV potential assessment on a street in Alesund, Norway (62.47°N, 6.15°E), showing (from left to right) original full view, rectified facade segment, semantic segmentation, LLM-suggested PV installation, and illustrative monthly/daily PV potential graphs.}
    \label{fig:alesund_workflow_detail}
\end{figure}

\begin{figure}
    \centering
    \includegraphics[width=\textwidth]{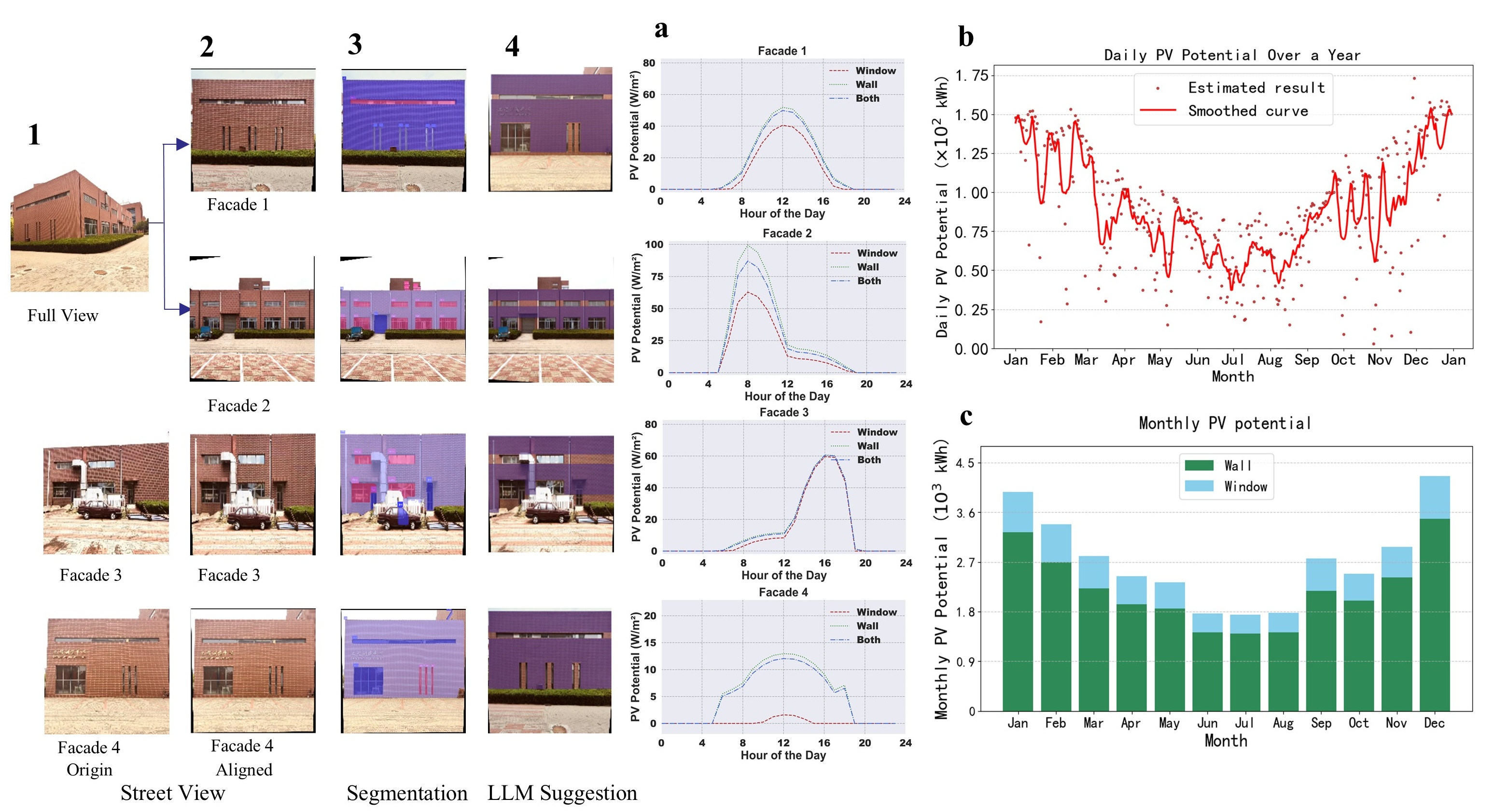} %
    \caption{Comparative analysis workflow illustration for facade PV potential assessment on a street in Tianjin, China (39.08°N, 117.20°E), showcasing similar stages as Figure \ref{fig:alesund_workflow_detail} and highlighting adaptability to different urban forms and climatic conditions.}
    \label{fig:tianjin_workflow_detail}
\end{figure}

The framework's applicability to semi-transparent window PV is also a relevant consideration. Although the default pipeline excludes windows, the framework is flexible enough to accommodate them by adjusting the LLM prompts. This allows for exploring scenarios that include window-based PV, considering factors like window-to-wall ratios and the energy-daylighting trade-off. Figures \ref{fig:alesund_workflow_detail} and \ref{fig:tianjin_workflow_detail} visualize the end-to-end workflow for specific streets, including window PV to demonstrate scalability. The numerical order of the images represents our step-by-step process flow of processing the raw data: acquiring the image from the building, image perspective correction, segmentation to obtain the desired part, and LLM interaction to give installation suggestions. Subfigure a summarizes the PV output of the façade for the three scenarios, b estimates the daily PV potential, and c shows the monthly potential. The analysis indicates that while both opaque façade and semi-transparent window installations show potential, the power generation per unit area for window PV is lower. However, for buildings with high window-to-wall ratios, the total contribution from window PV can be significant. As illustrated in the figures, geographic and climatic differences between Ålesund and Tianjin lead to significant variations in seasonal PV potential. In Ålesund, seasonal variations are particularly pronounced, with winter PV potential dropping significantly (e.g., 283.1 kWh in January) due to limited daylight and low sun angles. Conversely, potential peaks in spring, reaching 6,278 kWh in March and 7,079 kWh in April, driven by longer days and higher solar angles. In Tianjin, seasonal variations are less extreme but show a different pattern: potential is lowest in July (1,798 kWh) and highest in December (4,317 kWh).

\subsection{Effectiveness of Geometric Rectification}
\begin{figure}
    \centering
    \includegraphics[width=0.8\columnwidth]{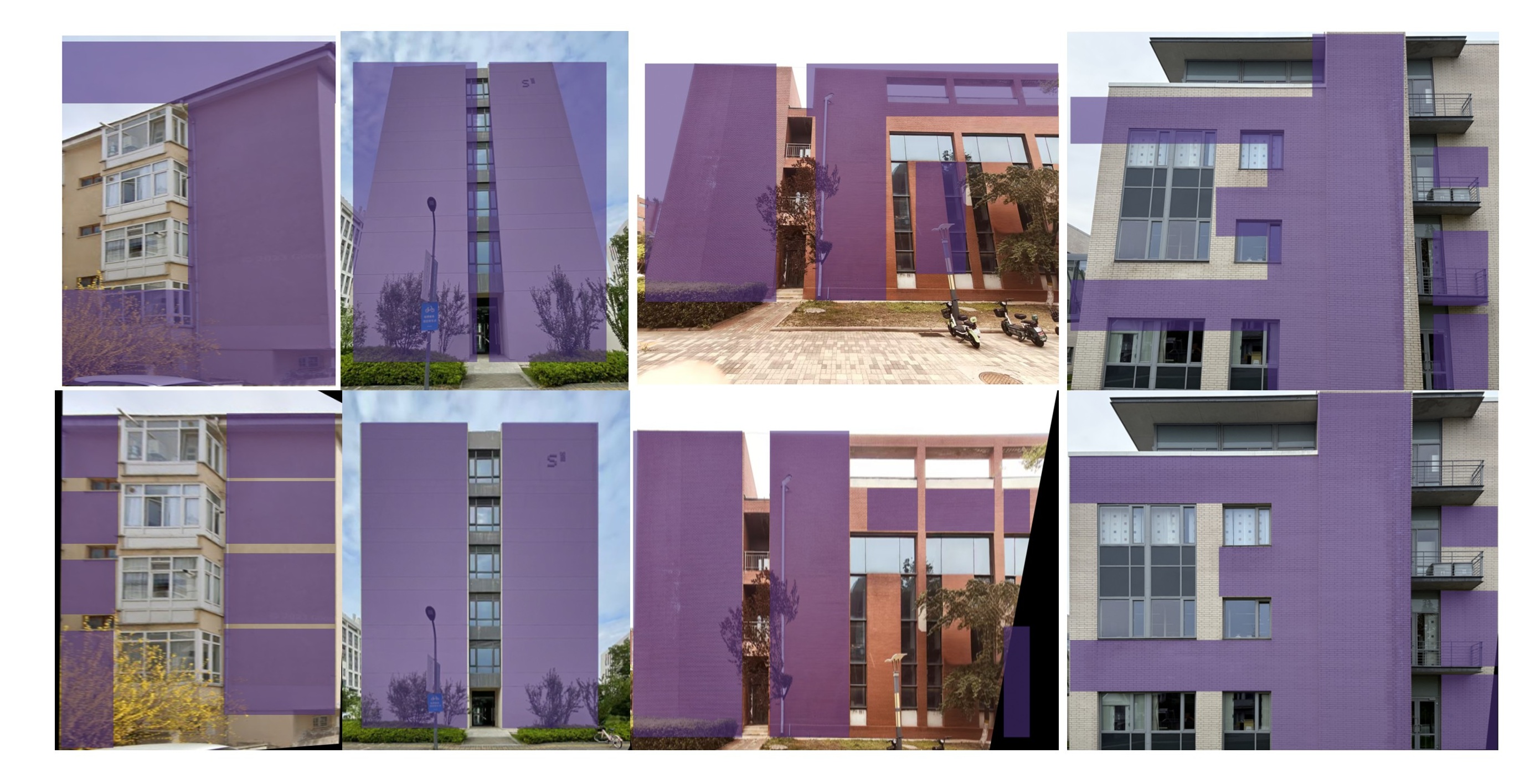} %
    \caption{Impact of Geometric Rectification on PV-Layout Generation Quality. Left: Processing an unrectified image leads to skewed masks and invalid PV layouts. Right: The rectified image yields clean semantic segmentation and accurate, installable PV layouts.}
    \label{fig:rectification_impact_comparison}
\end{figure}

Figure \ref{fig:rectification_impact_comparison} demonstrates the importance of geometric rectification by comparing PV layouts generated with and without this semantics-guided step. The rectified view aligns façade elements both vertically and horizontally, producing crisp, axis-aligned contours that preserve their true aspect ratios. This geometric regularity is crucial for the segmentation stage, as it provides the language model with accurately bounded wall zones. After rectification, the LLM positions PV modules flush against wall perimeters and correctly avoids non-installable areas like windows. In contrast, processing the unrectified image results in skewed masks and misplaced PV modules that either overlap windows or fail to cover usable wall surfaces. This comparison illustrates that geometric rectification provides a reliable foundation, enabling the downstream model to generate more accurate and practical PV layouts.


\subsection{Effectiveness of Vision-Model-based Segmentation}
The quality of semantic segmentation is paramount for accurate PV layout. Manual ground truth (GT) provides optimal segmentation but is impractical for large-scale application due to its high annotation cost. Fully automatic SAM operates without human input but produces non-semantic and often inaccurate masks that are unsuitable for direct use. Interactive SAM improves on the automatic version by using sparse user clicks, but it remains labor-intensive and prone to artifacts or omissions, especially near object boundaries. The automatic SAM often over- or under-segments regions, generating unusable noise. In contrast, our prompt-based LSA approach, guided by a single text instruction, consistently captures complete outlines of windows, doors, and walls without direct manual intervention. This superior mask quality, achievable without task-specific training, enables more logical placement of optimally sized PV modules, outperforming less precise segmentation methods. Table \ref{tab:Segmentation} compares these methods using the Mean Intersection over Union (mIoU) metric. The Mean Intersection over Union (mIoU) metric evaluates segmentation performance by quantifying the overlap between the predicted mask ($A$) and the ground-truth mask ($B$). mIoU averages the IoU scores across all classes, with higher scores indicating better accuracy. For a single class, IoU is defined as the ratio of the intersection area to the union area of the predicted and ground truth masks. mIoU extends this concept to multi-class segmentation. It is calculated by taking the average of the IoUs for each class, providing an overall measure of segmentation accuracy across all classes. A higher mIoU score indicates a better alignment between the model's predictions and the true segmentation.
\begin{table*}[ht]
\centering
\small
\caption{Comparative Analysis of Segmentation Paradigms.}
\label{tab:Segmentation}
\begin{tabular}{@{}c c c c@{}}
\toprule
     & \textbf{Human Intervention} & \textbf{Semantic Information} & \textbf{mIoU}\\ \hline
GT   & \multicolumn{1}{c}{$\surd$}      & \multicolumn{1}{c}{$\surd$}     & \multicolumn{1}{c}{1}     \\
Automatic SAM   & \multicolumn{1}{c}{$\times$}     & \multicolumn{1}{c}{$\times$}  & \multicolumn{1}{c}{0.753} \\
Interactive SAM & \multicolumn{1}{c}{$\surd$}       & \multicolumn{1}{c}{$\surd$}   & \multicolumn{1}{c}{0.787} \\
LSA      & \multicolumn{1}{c}{$\times$}         & \multicolumn{1}{c}{$\surd$}         & \multicolumn{1}{c}{0.814} \\
\bottomrule
\end{tabular}

\end{table*}

\subsection{Robustness of LLM-Driven PV Layout Reasoning}
\begin{figure}[t!]
    \centering
    \includegraphics[width=\columnwidth]{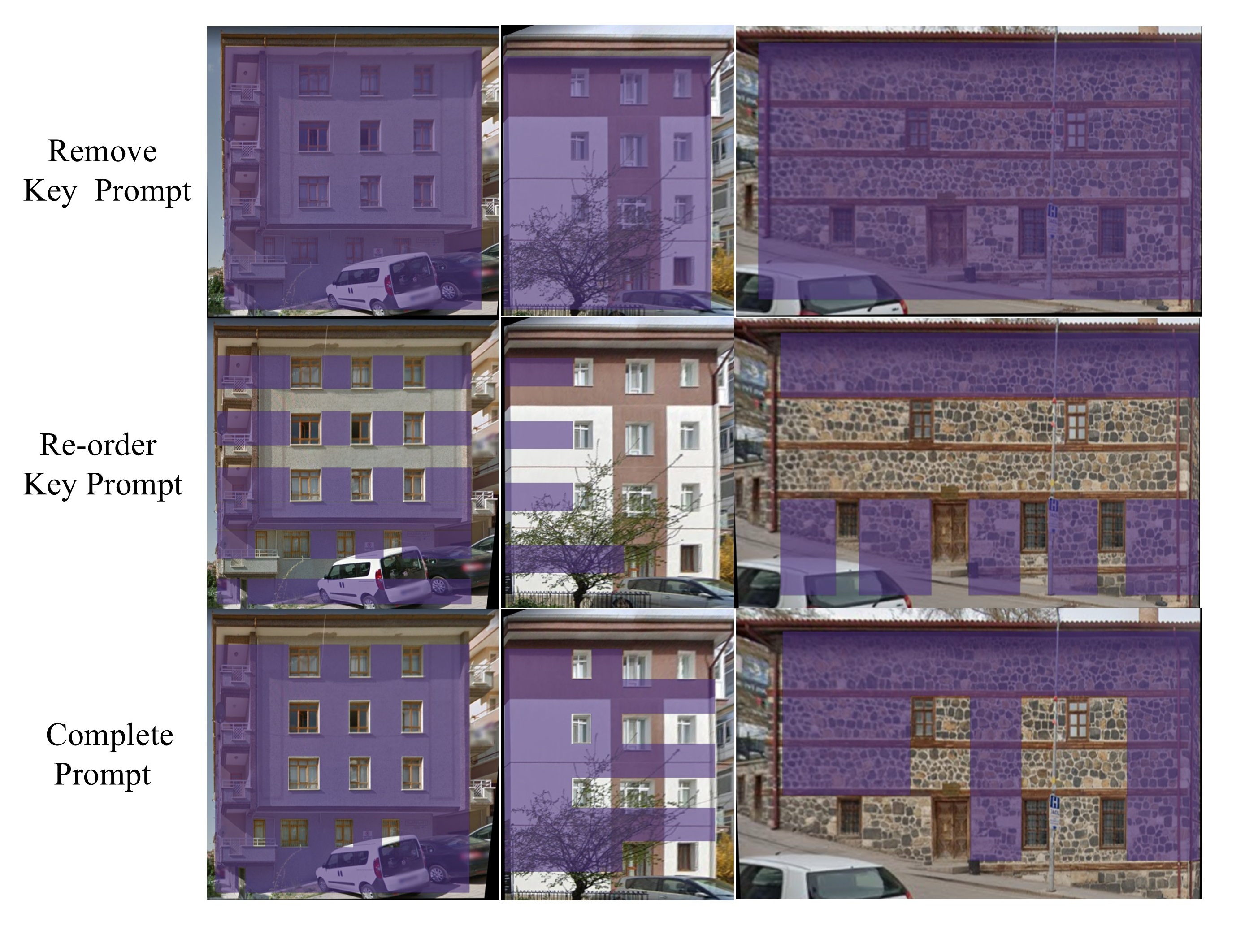} %
    \caption{Ablation study on LLM prompt engineering for PV Layout. (a) Result with key installation constraint sentences removed from the prompt: the LLM incorrectly places PV panels over windows. (b) Result with prompt clauses re-ordered (e.g., filter before partition): the LLM generates fragmented and suboptimal panel placements. (c) Result with the complete, structured prompt (as per Appendix): the LLM yields a physically reasonable, constrained, and optimized PV layout.}
    \label{fig:prompt_ablation_study}
\end{figure}

To evaluate the contribution and robustness of the LLM-driven reasoning module, we conducted two complementary studies.
First, we conducted a prompt-ablation test, systematically removing or reordering clauses within our engineered prompt template. As shown in Figure \ref{fig:prompt_ablation_study}, the results reveal the sensitivity and importance of the prompt structure. For instance, removing installation rules—such as prohibiting placement on windows or defining minimum dimensions—caused the LLM to suggest layouts over windows 
or create impractically small patches. Scrambling the logical order of the \textquotedblleft describe \textrightarrow partition\textquotedblright~steps resulted in fragmented and illogical layouts. These failures confirm that our engineered prompt effectively guides the LLM's spatial reasoning toward plausible and optimized solutions. Subsequently, an instruction sequence perturbation experiment examined the model's procedural robustness. Swapping the 'merge' and 'wall segmentation' operations significantly deteriorated output quality, resulting in fragmented layouts with discontinuous panel clusters. This sensitivity highlights the model's reliance on structured procedural guidance.
\begin{figure}[t!]
    \centering
    \includegraphics[width=0.8\columnwidth]{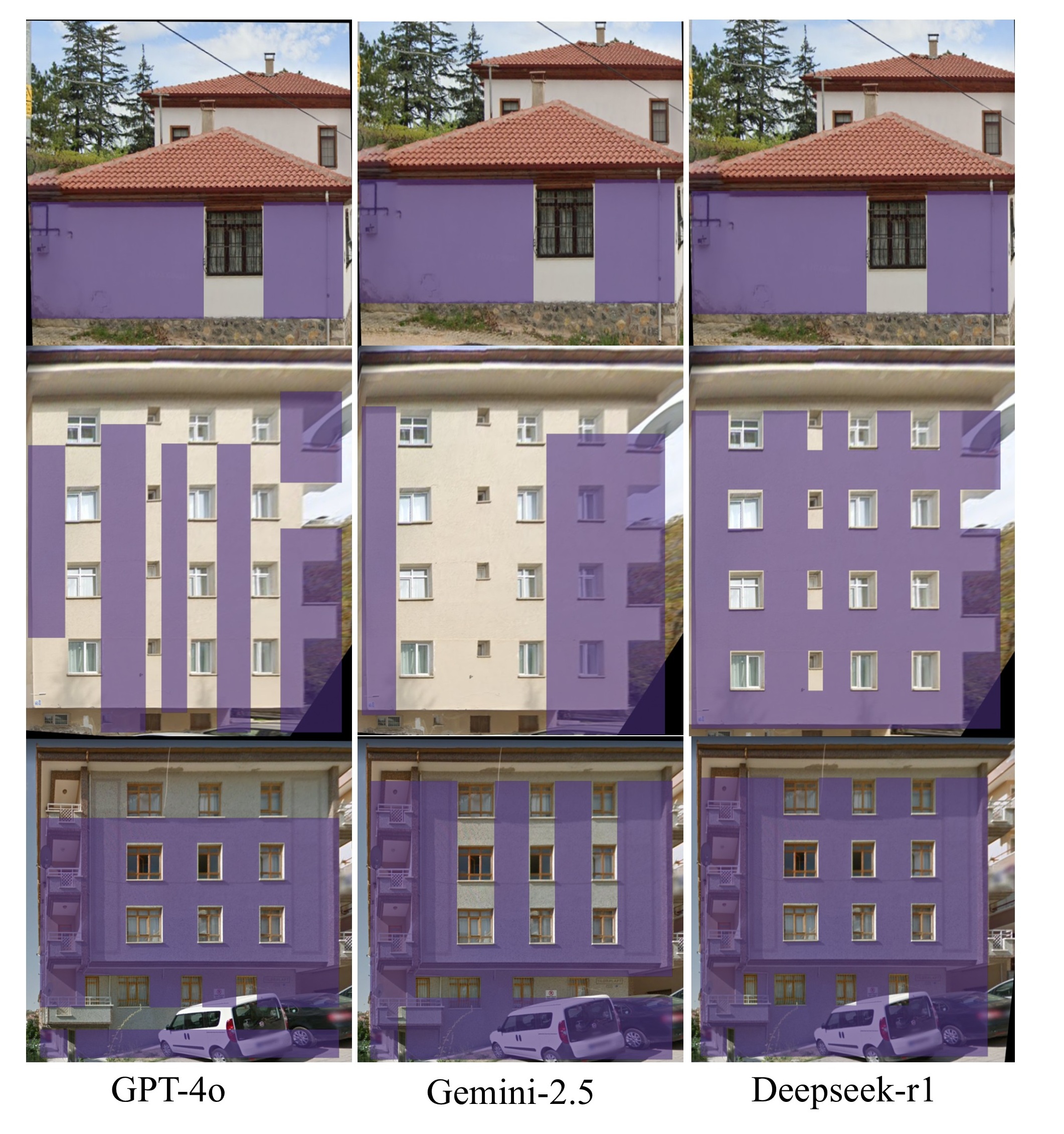} %
    \caption{Backbone LLM Comparison: Impact of Reasoning Capacity on PV-Layout Precision using the same input masks and prompt. (a) Layout generated by GPT-4o. (b) Layout generated by Gemini-2.5. (c) Layout generated by DeepSeek-r1. Differences in layout coherence, alignment, and utilization of space are visible, highlighting that stronger reasoning abilities contribute to more precise PV layouts.}
    \label{fig:model_comparison_study}
\end{figure}

Second, we evaluated the impact of the LLM backbone by swapping the model while keeping the input data and prompt template constant. We compared GPT-4o (our default), Gemini-2.5, and DeepSeek-r1. As shown in Figure \ref{fig:model_comparison_study}, DeepSeek-r1, a strong open-source alternative, generally produced highly coherent and well-aligned layouts. GPT-4o also performed well, generating slightly more conservative regions but with excellent adherence to constraints. In contrast, Gemini-2.5 occasionally misinterpreted imperfect segmentation masks—for example, by merging adjacent windows—which led to less precise layouts. This performance variation suggests that while the prompt is model-agnostic, LLMs with stronger instruction-following and spatial reasoning capabilities yield more precise and reliable results.

\begin{table}[]
\caption{Comparison of LLM backbone performance for spatial reasoning tasks. Mean Request Count indicates the average number of API calls required to obtain syntactically valid JSON output due to LLM response formatting inconsistencies.}
{\tiny
\resizebox{\columnwidth}{!}{
\begin{tabular}{cccc}
\toprule
LLM Backbone & Average Inference time(s) & Mean Request Count & Estimation Error(\%)\\
\midrule
GPT-4o       & 31.3                      & 3.1                & 6.7                \\
Gemini-2.5   & 36.9                      & 2.9                & 9.4                 \\
Deepseek-r1  & 81.7                      & 4.7                & 5.2                \\
\bottomrule
\end{tabular}%
}}
\end{table}

The Mean Request Count metric reveals an important practical consideration: LLMs occasionally produce malformed JSON responses that require re-querying. GPT-4o and Gemini-2.5 demonstrate superior response formatting consistency, while Deepseek-r1 requires more retry attempts but achieves the lowest estimation error when valid responses are obtained. This highlights the trade-off between response reliability and accuracy in structured data generation tasks.

In summary, these experiments demonstrate that the engineered prompt framework is crucial for generating physically plausible and optimized PV layouts. They also show that while the framework is flexible, the choice of the underlying LLM influences the final design quality.

\subsection{Energy Yield Estimation and Regional Sensitivity}
\label{sec:energy_yield_analysis}
Following the validation of installable façade areas, the final step is to estimate the potential energy yield (kWh) under realistic conditions. This analysis demonstrates the practical utility of the framework's geometric and semantic outputs for energy modeling. We compared ideal clear-sky simulations (representing an upper bound) with realistic simulations using Typical Meteorological Year (TMY) weather data. TMY data, which includes factors like cloud cover and temperature, provides a more realistic estimate of annual energy production.

\begin{table}[]
\caption{PV Potential Comparison by Building Type (Aggregated from Sample Set, TMY-based Yield)}
\label{tab:building_type_comparison}
\resizebox{\columnwidth}{!}{%
\begin{tabular}{ccccc}
\toprule
\textbf{\begin{tabular}[c]{@{}c@{}}Building \ Type
\end{tabular}} & 
  \textbf{\begin{tabular}[c]{@{}c@{}}Building\  Count
\end{tabular}} & 
  \textbf{\begin{tabular}[c]{@{}c@{}}Total Availuable \ Installation area(m$^2$)
\end{tabular}} & 
  \textbf{\begin{tabular}[c]{@{}c@{}}Ideal Clear-Sky\ Power Gen. (MWh/yr)
\end{tabular}} & 
  \textbf{\begin{tabular}[c]{@{}c@{}}Actual TMY-based\ Power Gen. (MWh/yr)
\end{tabular}} \\ \midrule
Low-rise building(s)  & 20 & 3031.6   & 471.331  & 398.645  \\
Mid-rise building(s)  & 40 & 14495.01 & 2117.891 & 1541.666 \\
High-rise building(s) & 20 & 18275.87 & 2565.478 & 1503.826 \\
\bottomrule
\end{tabular}% 
}
\end{table}
\begin{table}[]
\caption{Regional Comparison of Building-Integrated PV Potential (Aggregated from Sample Set)}
\label{tab:regional_comparison_expanded}
\resizebox{\columnwidth}{!}{%
\begin{tabular}{ccccc}
\toprule
\textbf{Location} & \textbf{\begin{tabular}[c]{@{}c@{}}Building \ Count
\end{tabular}} & 
  \textbf{\begin{tabular}[c]{@{}c@{}}Total Available \ Installation Area (m$^2$)
\end{tabular}} & 
  \textbf{\begin{tabular}[c]{@{}c@{}}Ideal Clear-Sky \ Power Gen. (MWh/yr)
\end{tabular}} & 
  \textbf{\begin{tabular}[c]{@{}c@{}}Actual TMY-based \ Power Gen. (MWh/yr)
\end{tabular}} \\ \midrule
Hangzhou, China & 20 & 11942.68 & 2150.073 & 976.646  \\
Tianjin, China  & 20 & 13631.73 & 1414.299 & 1119.715 \\
Ankara, Turkey  & 20 & 5556.76  & 895.559  & 712.900  \\
Ålesund, Norway & 20 & 4671.31  & 694.806  & 634.876  \\
\bottomrule
\end{tabular}% 
}
\end{table}
The analysis considered both building typology and geographical location. First, buildings were categorized as low-, mid-, and high-rise, and their aggregated generation potential was calculated using TMY data (Table \ref{tab:building_type_comparison}). The results indicate that buildings of all heights possess significant façade PV potential. As building height increases, so does the available façade area. Consequently, high-rise buildings show greater absolute generation potential under similar conditions. However, the collective potential of low- and mid-rise buildings, which are more numerous in cities, should not be overlooked.

\begin{table}[]
\caption{Comparison of Façade and Rooftop PV Generation Across Different Building Types (Equal Area Basis).}
\label{tab:Comparison-1}
\resizebox{\columnwidth}{!}{%
\begin{tabular}{ccccc}
\toprule
\textbf{Building Type} &
  \textbf{\begin{tabular}[c]{@{}c@{}}Total Available \ Installation area(m$^2$)
\end{tabular}} &
  \textbf{\begin{tabular}[c]{@{}c@{}}Facade Power \ Gen. (MWh/yr)
\end{tabular}} &
  \textbf{\begin{tabular}[c]{@{}c@{}}Rooftop Power \ Gen. (MWh/yr)
\end{tabular}} &
  \textbf{\begin{tabular}[c]{@{}c@{}}Facade-to-Roof \ Generation Ratio (\%)
\end{tabular}} 
\\
\midrule
Low-rise building(s)  & 3031.6   & 398.645  & 783.979  & 50.8 \\
Mid-rise building(s)  & 14495.01 & 1541.666 & 3404.659 & 45.3 \\
High-rise building(s) & 18275.87 & 1503.826 & 4111.951 & 36.5 \\
\bottomrule
\end{tabular}% 
}
\end{table}

Next, we analyzed the generation potential by geographical region, revealing unique characteristics governed by local climate (Table \ref{tab:regional_comparison_expanded}). Notably, while the Hangzhou building samples have a larger installable area and higher theoretical (clear-sky) yield, their actual TMY-based energy generation is lower than that of the Tianjin samples. This discrepancy arises because Hangzhou experiences more overcast weather, which reduces effective sunlight hours and lowers the total energy output. Conversely, Tianjin, despite potentially less raw area in this sample, benefits from a sunnier climate, leading to higher actual generation. This finding underscores the importance of using local meteorological data for realistic PV potential assessments, a key feature of our framework.

\begin{table}[]
\caption{Comparison of Façade and Rooftop PV Power Generation of Buildings in Different Regions Under Equal Area Conditions.}
\label{tab:Comparison-2}
\resizebox{\columnwidth}{!}{%
\begin{tabular}{ccccc}
\toprule
\textbf{Location} &
  \textbf{\begin{tabular}[c]{@{}c@{}}Total Available \ Installation Area (m$^2$)
\end{tabular}} &
  \textbf{\begin{tabular}[c]{@{}c@{}}Facade Power\ Gen. (MWh/yr)
\end{tabular}} &
  \textbf{\begin{tabular}[c]{@{}c@{}}Rooftop Power\ Gen. (MWh/yr)
\end{tabular}} &
  \textbf{\begin{tabular}[c]{@{}c@{}}Facade-to-Roof\ Gen. Ratio (\%)
\end{tabular}} \\
\midrule
Hangzhou, China & 11942.68 & 976.646  & 2474.978 & 39.5 \\
Tianjin, China  & 13631.73 & 1119.715 & 3190.539 & 35.1 \\
Ankara, Turkey  & 5556.76  & 712.900  & 1384.276 & 51.5 \\
Ålesund, Norway & 4671.31  & 634.876  & 1250.795 & 49.2 \\
\bottomrule
\end{tabular}% 
}
\end{table}

Finally, we compared the energy yield of façade PV with that of traditional rooftop PV. Tables \ref{tab:Comparison-1} and \ref{tab:Comparison-2} present results on both an equal-area basis and a total-available-area basis. For equal installation areas, façade PV yields 35.1\% to 51.5\% of rooftop PV energy. However, this comparison understates façade potential in high-rise contexts.

\textbf{Total Area Analysis:} Our building sample analysis reveals that façade area significantly exceeds rooftop area, particularly for high-rise buildings. For buildings >10 stories, the total façade area averages 3.2× the rooftop area; for buildings >20 stories, this ratio increases to 4.8×. When comparing total potential energy generation (façade area × façade yield vs. rooftop area × rooftop yield), façade PV achieves 1.1× to 2.5× the total energy output of rooftop PV for high-rise buildings. This analysis demonstrates that despite lower per-unit-area efficiency, façades represent the dominant renewable energy resource for tall urban buildings, making them critical for comprehensive urban energy planning.

\section{Limitations and Failure Modes}
\label{sec:limitations}

While the SF-SPA framework demonstrates promising performance, several limitations and potential failure modes must be acknowledged for transparent evaluation and future improvement.

\subsection{Geometric and 3D Limitations}
The homographic rectification assumes perfectly planar facades, which breaks down for: (1) curved or non-rectilinear building geometries, (2) facades with significant depth variations (deep-set windows, protruding balconies), and (3) complex architectural features. Our method is most suitable for modern commercial and residential buildings with relatively flat facades. The 2D approach fundamentally cannot account for inter-building shading, which is critical in dense urban environments. We estimate this limitation introduces 10-25\% overestimation of energy yield in high-density areas.

\subsection{Rectification Failure Modes}
The geometric rectification requires at least four clearly visible, unoccluded rectangular features (typically windows) to establish the homography. Failure scenarios include: (1) facades with no suitable rectangular features, (2) heavily occluded views due to vegetation or adjacent structures, (3) facades with irregular window arrangements, and (4) images captured at extreme viewing angles (>60° from facade normal). In our dataset, 8\% of initial candidates were excluded due to rectification failures.

\subsection{LLM Consistency and Latency}
LLM-based spatial reasoning exhibits inherent non-determinism. Testing GPT-4 with identical inputs across 10 runs showed area estimation variance of ±4.2\% (standard deviation). This variability stems from the model's probabilistic nature and sensitivity to prompt variations. The current LLM processing time (97 seconds average) represents a significant bottleneck for real-time applications. Additionally, LLM performance degrades for highly complex facades with >20 distinct architectural elements.

\subsection{Segmentation Robustness}
The vision-language segmentation may fail for: (1) unconventional architectural styles not well-represented in training data, (2) facades with poor lighting conditions or low image quality, (3) buildings with highly reflective surfaces causing glare artifacts, and (4) facades with complex material textures that confuse semantic boundaries. Historical or vernacular architecture poses particular challenges due to irregular geometries and non-standard elements.

\subsection{Scale and Deployment Considerations}
The manual scaling step remains a critical bottleneck, introducing potential errors and limiting full automation. Weather conditions, seasonal variations, and long-term facade degradation are not considered in the current energy modeling. The framework's accuracy may vary across different geographic regions due to architectural style variations and local building codes not captured in the training data.

\section{Conclusions}
\label{sec:conclusions}
This study presented SF-SPA, a novel automated framework that integrates vision foundation models and LLMs to assess solar PV potential on urban façades. By combining geometric rectification, zero-shot segmentation, and LLM-driven reasoning, our framework effectively extracts installable PV areas from single street-view images, overcoming the limitations of traditional methods. Validation on a diverse dataset demonstrated high accuracy, with an average area error of 6.2\% ± 2.8\%, and significant efficiency at approximately 100 seconds per building. The energy yield estimations, which incorporate meteorological data, highlight the framework's practical utility for regional potential studies and planning for Building-Integrated Photovoltaics (BIPV).

\textbf{Policy and Economic Implications:} This automated assessment capability has significant implications for urban energy policy and economic planning. The framework enables rapid, city-scale screening of facade PV potential, supporting evidence-based policy development for renewable energy targets and building codes. For urban planners, the tool provides quantitative data to prioritize districts for BIPV incentives and infrastructure investments. Energy companies can use the framework for market assessment and project feasibility studies, while building owners gain access to preliminary PV potential estimates without expensive consultancy fees. The scalability of the approach makes it particularly valuable for developing countries where manual assessment resources are limited but renewable energy deployment is critical.

Key limitations include the reliance on 2D images and manual scaling requirements. Future work should enhance geometric fidelity using 3D models to better analyze complex façades and inter-building shading. Further research should improve the robustness of the rectification process, automate the scaling step, and account for dynamic factors such as component soiling. Integrating economic analysis and policy factors will advance the framework toward a holistic decision-support tool for BIPV adoption. The framework's modular design facilitates future advancements as AI technologies evolve, paving the way for more comprehensive and data-driven urban energy solutions.
\section*{Acknowledgement}
The work is supported by the Marie Skłodowska-Curie Postdoctoral Individual Fellowship under Grant No. 101154277.

\section*{Appendix: LLM Prompt Template}
The following is a representative template for the prompt used to guide the LLM in Step 3 (Partition usable wall) and Step 4 (Qualify for PV installation) of the PV Layout Reasoning stage (Section \ref{sec:llm_reasoning}). Placeholders like {w\_m} are filled programmatically.

\noindent\textbf{Role}: You are an expert civil engineer specializing in Building-Integrated Photovoltaics (BIPV). Your task is to determine the optimal layout for PV panels on a building facade based on provided geometric and semantic data. You must strictly follow all instructions and output formats.

\vspace{1em}
\noindent\textbf{Task}: Given the dimensions of a building facade and the locations of obstructions (windows, doors, balconies, etc.), identify all possible rectangular areas suitable for PV installation.

\vspace{1em}
\noindent\textbf{Step 1: Understand Metric Canvas (Provided for Context)}\\The building facade image has been rectified to a front-on view. The relevant parameters are:
\begin{itemize}
    \item Real-world dimensions: width $= \{w_m\}$ meters, height $= \{h_m\}$ meters.
    \item Pixel dimensions: width $= \{w_{px}\}$ pixels, height $= \{h_{px}\}$ pixels.
\end{itemize}
All following coordinates are in pixels, with the origin $(0, 0)$ at the top-left corner of the rectified image.

\vspace{1em}
\noindent\textbf{Step 2: Identify Obstructions (Provided Semantic Layout)}\\The following are obstructions on the facade where PV panels \textbf{cannot} be installed. They are provided as lists of bounding boxes, each defined by $[x_1, y_1, x_2, y_2]$, representing the top-left $(x_1, y_1)$ and bottom-right $(x_2, y_2)$ pixel coordinates:
\begin{itemize}
    \item Windows: \{\texttt{list\_of\_window\_boxes}\}
    \item Doors: \{\texttt{list\_of\_door\_boxes}\}
    \item Balconies: \{\texttt{list\_of\_balcony\_boxes}\}
    \item Other obstructions: \{\texttt{list\_of\_other\_boxes}\}
\end{itemize}
\textit{Note: If a list is empty, it means no such obstructions were detected.}

\vspace{1em}
\noindent\textbf{Step 3: Partition Usable Wall Area}\\Your primary goal is to analyze the remaining \textquotedblleft wall\textquotedblright~area (i.e., total facade area minus all areas occupied by the obstructions from Step 2). You must subdivide this available wall space into a set of non-overlapping rectangular regions suitable for potential PV installation. The following critical rules apply:
\begin{enumerate}[label=(\alph*)]
    \item \textbf{Maximize Area per Region:} Each identified rectangular region should be as large as possible. Minimize the total number of rectangles by favoring larger, contiguous PV arrays.
    \item \textbf{Mutual Exclusivity and Obstruction Avoidance:} The generated rectangular regions must not overlap with any of the obstruction bounding boxes from Step 2, nor with each other.
    \item \textbf{Comprehensive Coverage:} The set of identified rectangular regions should aim to cover as much of the available, unobstructed wall surface as possible.
    \item \textbf{Merging for Optimization:} If merging two or more adjacent, smaller valid sub-regions (that individually satisfy all rules) results in a larger valid rectangular region without violating exclusivity or obstruction rules, this merge operation should be performed.
\end{enumerate}

\vspace{1em}
\noindent\textbf{Step 4: Qualify Rectangles for PV Installation}\\From the list of potential wall rectangles generated in Step 3, filter them based on practical PV module installation constraints. A rectangle is considered suitable for PV installation \textbf{only if} it meets \textbf{both} of the following dimensional criteria (converted to real-world meters using the scale factor from Step 1):
\begin{itemize}
    \item The shorter side (width or height) is at least \{min\_short\_edge\_m\} meters.
    \item The longer side (width or height) is at least \{min\_long\_edge\_m\} meters.
\end{itemize}
Rectangles failing to meet either criterion must be discarded.

\vspace{1em}
\noindent\textbf{Output Format:}\\Provide your final answer strictly as a \texttt{JSON} object. This object should have a single key named \texttt{\"installable\_rectangles\"}, whose value is a list of valid rectangular areas that passed all criteria in Step 4. Each rectangle in the list must be represented by its pixel coordinates in the format $[x_1, y_1, x_2, y_2]$. If no rectangles are found to be suitable, return an empty list for \texttt{\"installable\_rectangles\".}

\textit{Example of a valid output:}
\begin{verbatim}
{
  "installable_rectangles": [
    [150, 50, 300, 400],
    [550, 50, 700, 400],
    [50, 450, 700, 600]
  ]
}
\end{verbatim}

\textit{Example if no suitable areas are found:}
\begin{verbatim}
{
  "installable_rectangles": []
}
\end{verbatim}

\noindent\textbf{Begin your analysis now.}
\appendix
\section{More Implementation Details}
\subsection{Complete Prompt}
\begin{itemize}
\item \textbf{Task:} You are an intelligent assistant for recommending solar panel installations. Your task is to analyze building information step by step and provide suitable locations for the placement of solar panels on building surfaces.
\item \textbf{Step 1:} Obtain information about the image described by the user (describe the image composition and location information in text), which, usually, contains walls, roofs, doors, windows, etc. The roof, door, window, etc. are part of the wall.
Example of user input:
- Building surface description size and actual size:
        \begin{itemize}
	\item The size of the building is described as: (1200, 800), where the x-direction is 1200 and the y-direction is 800.
	\item The actual size of the corresponding building surface is 12 meters, 6 meters.
	\item The scale is: x-direction: 12 meters/1200, y-direction: 6 meters/800.
        \item There are the following objects in the building surface: 'wall', ‘window’, 'door', the corresponding position information of these objects is shown below:
	\item Take the upper left corner as the (0,0) point, to the right is the x-axis positive direction, and down is the y-axis positive direction.
	\item When describing each object, the meaning of the position information is $[x_{min},y_{min},x_{max},y_{max}]$.
	\item 'wall': the wall contains the position of the window and the door, its specific position is: wall1: $[0,0,1200,800]$.
	\item 'window': the wall contains multiple windows, location information is: window1: $[50,200,250,400]$, window2: $[300,200,450,400]$, window3: $[850,100,1150,500]$.
	\item 'door': there is a door in the wall, location information is: door 1: $[600,520,750,800]$.
        \end{itemize}
\item \textbf{Step 2:} Divide the entire wall into sub-parts according to the following rules: 
    \begin{itemize}
        \item  Each roof, each window, each door, etc. must be divided into a separate part
        \item The remaining part of the wall, i.e., the location that does not contain the roof, doors, windows, etc., is then divided according to the following rules:
        \begin{itemize}
            \item After the division of the area, as far as possible, a single area of large, small number of areas.
            \item Each part of the partitioned area should be mutually exclusive and contain the entire wall surface except for the components.
            \item According to the result of the partition, merge the areas that can be merged, for example: $[0,0,100,150]$ and $[0,150,100,450]$ can be merged into $[0,0,100,450]$.
        \end{itemize}
    \end{itemize}
\item \textbf{step 3:} Considering the size and type of each area after division, determine whether the area is suitable for the installation of solar panels. The suitable part of the installation must meet the following requirements: \begin{itemize}
        \item Only the wall part can be installed, i.e. the roof, doors, windows and other locations cannot be installed with solar panels.
        \item The short side of the area must be greater than or equal to 1 meters, and the long side of the area must be greater than or equal to 1.2 meters.
    \end{itemize}
\item \textbf{step 4:} Give the area suitable for installation, in the following example format, using '\#' to comment if there are comments: \par
[   \par
    \hspace{2em} $[x1_{min},y1_{min},x1_{max},y1_{max}]$, \par
    \hspace{5em}\t...    \par
    \hspace{2em}
    $[xn_{min},yn_{min},xn_{max},yn_{max}]$ \par
]
\end{itemize}
\begin{itemize}
\item \textbf{Task:} Next is a specific scenario, following the steps required by the above example and the specific details of each step, analyze the new scenario specifics and give recommendations for the installation of solar PV panels on the wall of a building.
\item The described and actual size of the building surface is: (X, Y) where the x-direction is X and the y-direction is Y.
\item The actual size of the corresponding background is M meters in the x-direction and N meters in the y-direction.
\item There are the following objects on the surface of the building: 'wall', ‘window’, 'door', and the corresponding positions of these objects are as follows:
    \begin{itemize}
        \item Take the upper left corner as the (0,0) point, to the right is the x-axis positive direction, and down is the y-axis positive direction.
        \item When describing each object, the meaning of the specific position is $[x_{min},y_{min},x_{max},y_{max}]$.
        \item 'wall': the wall contains the window and door position, its specific position is: wall1: $[x1_{min},y1_{min},x1_{max},y1_{max}]$, wall2:$[x2_{min},y2_{min},$ \par $x2_{max},y2_{max}]$, etc.
        \item 'window': the wall has more than one window, the location information is: window1: $[x1_{min},y1_{min},x1_{max},y1_{max}]$, etc.
        \item there is a door on the wall, the location information is: door1: $[x1_{min},y1_{min},x1_{max},y1_{max}]$, etc.
    \end{itemize}
\end{itemize}
\label{app1}
\subsection{Environment Setting}
Experiments were conducted on an NVIDIA RTX 3090 GPU. As our framework is training-free, most modern graphics cards are sufficient for this task. Interactions with the Large Language Models (LLMs) were conducted via their respective APIs using a multi-round conversational format to ensure consistency. Key hyperparameters were fixed for all models: the temperature was set to 0.0 for deterministic outputs, and the maximum token limit was set to 1000 to prevent truncation.

\bibliographystyle{elsarticle-num}
% \bibliography{references} 

\end{document}